\documentclass{article} 
\usepackage{nips14submit_e,times}
\usepackage{hyperref}
\usepackage{url}

\usepackage[toc,page]{appendix}
\usepackage{times,amsmath,epsfig}
\usepackage{epstopdf}
\usepackage{graphicx}
\usepackage{url}
\usepackage{subfigure}
\usepackage{amssymb,amsfonts,verbatim}
\usepackage{amsthm}
\usepackage{bm}
\usepackage{mathtools}
\usepackage{multirow}
\usepackage{graphicx}
\usepackage{subfigure}
\usepackage{algorithm}
\usepackage{algorithmic}
\usepackage{xcolor}
\usepackage{adjustbox}[2011/08/07]
\usepackage{xspace}
\usepackage{booktabs} 
\newcommand{\sizetraining}{0.30}

\title{Group $K$-Means}

\author{
\begin{tabular}[t]{ccc}
\vspace{0.1cm}
\rule{0pt}{12pt}Jianfeng Wang$^{\dagger}$ & \rule{0pt}{12pt}Shuicheng Yan$^{\ddagger}$ & 
\rule{0pt}{12pt}Yi Yang$^{\star}$   \\
Mohan S Kankanhalli$^{\ddagger}$ & 
Shipeng Li$^{\sharp}$ & Jingdong Wang$^{\sharp}$ \\
& & \\
\multicolumn{3}{c}{\textnormal{$^{\dagger}$University of Science and Technology of China}} \\
\multicolumn{3}{c}{\textnormal{$^{\ddagger}$National University of Singapore}} \\
\multicolumn{3}{c}{\textnormal{$^{\star}$ The University of Queensland}} \\
\multicolumn{3}{c}{\textnormal{$^{\sharp}$Microsoft Research, Beijing}} \\
\end{tabular}
}
\nipsfinalcopy 

\begin{document}

\maketitle

\begin{abstract}
We study how
to learn multiple dictionaries
from a dataset, and 
approximate any data point by the sum of 
the codewords each chosen from the corresponding dictionary.
Although theoretically low approximation errors
can be achieved by the global solution, an effective solution has not been well studied in practice. 
To solve the problem, we propose a simple yet effective 
algorithm \textit{Group $K$-Means}.
Specifically,
we take each dictionary, or any 
two selected dictionaries,
as a group of $K$-means cluster centers, and then deal with the approximation issue by minimizing the approximation errors. Besides, we propose a hierarchical initialization for such a non-convex problem.
Experimental results well validate the effectiveness of the approach.
\end{abstract}

\section{Introduction}
$K$-means is a well-known clustering algorithm and 
has been 
widely applied in numerous applications. 
The algorithm aims to partition $N$ $P$-dimensional points into 
$K$ clusters in which each point belongs to the cluster with the nearest mean.
Let 
$\mathcal{X} = \{\mathbf{x}_1, \cdots, \mathbf{x}_N\} \subset \mathbb{R}^{P}$ 
be the dataset and 
$\mathcal{D} = \{\mathbf{d}_1, 
\cdots, \mathbf{d}_K\} \subset \mathbb{R}^{P}$ 
be the cluster centers. 
The clusters $\mathcal{D}$
are learned by minimizing
\begin{align}
\sum_{\mathbf{x}\in \mathcal{X}}\min_{k}{\|\mathbf{x} - \mathbf{d}_k\|_2^2},
\end{align}
where $\|\cdot\|_p$ denotes the $l_p$ norm.
The objective function
can be iteratively minimized~\cite{Lloyd82}.
Each iteration involves an
\textit{assignment step} and an \textit{update step}.
In the former step, the nearest center of each point
is calculated, while the latter computes the mean
of the points assigned into the same cluster. 
Each point can be represented by the index of the nearest cluster center, which requires $\lceil \log_{2}{K} \rceil$
\footnote{In the following, without confusion we omit the $\lceil z \rceil$ operator,
which represents the smallest integer that is not smaller than $z$. 
}
bits.

The assignment of each point requires $O(KP)$ time cost. 
When the number of clusters is huge, it is prohibitive to perform 
the exact $K$-means due to the high time cost. 
To solve the scalability issue, we can split the $P$-dimensional vector
into $M$ subvectors as 
in~\cite{GeHK013, JegouDS11, NorouziF13}.
Then the standard $K$-means algorithm is applied on each subvector, 
resulting in $K^{M}$ cluster centers but with $O(KP)$ assignment time cost.
The number of bits required to represent each point is $\log_2{K^M} = M\log_{2}{K}$.

Recently, multiple dictionaries are proposed in~\cite{WangWSXSL14, DuW14, BabenkoL14}.
Each dictionary contributes one codeword and
the summation of these codewords is used to approximate one data point.
Let $\mathcal{D}^{c} = \{\mathbf{d}^c_1, \cdots, \mathbf{d}^c_K\} \subset \mathbb{R}^{P}$
be the $c$-th dictionary with $c \in \{1, \cdots, C\}$.
The objective is to minimize 
\begin{align}
\sum_{\mathbf{x} \in \mathcal{X}}\min_{k_1, \cdots, k_C}\left\|\mathbf{x} - \sum_{c}\mathbf{d}^{c}_{k_c}\right\|_2^2
\label{eqn:gkmeans_obj}
\end{align}
to learn the dictionaries $\{\mathbf{D}^{c}\}_{c = 1}^{C}$.
In~\cite{WangWSXSL14}, the problem is studied on the subvector, and 
can be seen as a special case where the subvector is equal to the full vector
or the number of subvectors is $1$.
This problem is also explored in~\cite{ZhangDW14}, but with an additional constraint to
make the scheme more suitable for Euclidean approximate nearest neighbor search.
To represent each point, we need $C\log_2{K}$ bits to indicate which codewords are selected from all the dictionaries.

It is easily verified based on~\cite{WangWSXSL14} that a global optimal solution can give a lower  distortion error
than~\cite{JegouDS11, NorouziF13}
under the same code lengths. 
However, it is very challenging to obtain the global solution due to the non-convexity of the problem. 
In~\cite{WangWSXSL14}, an intuitive recursive algorithm is proposed, but
the complexity is exponential with the number of dictionaries, and it is
less scalable with a larger $C$. 

In this paper, 
we propose a simple yet effective algorithm, named \textit{Group $K$-Means} (shorted as gk-means)\footnote{The algorithm is 
similar with~\cite{Barnes94}, but we conduct the 
research independently and apply it in data/feature compression and image retrieval. }, 
to solve the problem.
Specifically, we take each dictionary or any two consecutive dictionaries as a group of 
$K$-means
clusters,
and solve the assignment with a linear time complexity to the number of dictionaries. 
Due to the non-convexity of the problem, we propose a hierarchical 
scheme composed of multiple stages to
initialize the dictionaries. 
Each stage solves a subproblem 
where a portion of the entries in the 
dictionaries are
enforced to be $0$.
Experimental results have verified the effectiveness of our approach. 

\section{Group $K$-means}

To minimize Eqn.~(\ref{eqn:gkmeans_obj}), we 
iteratively perform the assignment step
and the update step as shown in Alg.~\ref{alg:overall}, 
and introduced in the first two subsections. 
The former computes the assignments $(k_1, \cdots, k_C)$ of each point 
$\mathbf{x}$
based on
the dictionaries and the assignments in the previous iteration. 
The latter updates the multiple dictionaries based on the current assignments.
Besides, the initialization of the dictionaries and the assignments
is introduced in the last subsection.

\begin{algorithm}[t]
\caption{Iterative optimization in gk-means}
\label{alg:overall}
\begin{algorithmic}[1]
	\REQUIRE
		Dataset	$\mathcal{X} = \{\mathbf{x}_1, \cdots, \mathbf{x}_N\}$,
		number of dictionaries $C$
	\ENSURE
		Multiple dictionaries $\{\mathcal{D}^{c}, c\in \{1, \cdots, C\}\}$
	\STATE
		Initialize multiple dictionaries $\{\mathcal{D}^{c},  c\in \{1, \cdots, C\}\}$
		by Sec.~\ref{sec:init_dictionary}
	\STATE
		Initialize the assignments $\{k_1, \cdots, k_C\}$
		for each point $\mathbf{x} \in \mathcal{X}$ by Sec.~\ref{sec:init_assignment}
	\WHILE {!converged}
		\STATE
			Update multiple dictionaries by Sec.~\ref{sec:update}
		\STATE
			Compute assignments by Sec.~\ref{sec:assignment}
			\label{line:assignment}
	\ENDWHILE
\end{algorithmic}
\end{algorithm}

\subsection{Assignment Step}
\label{sec:assignment}

\subsubsection{Order-1 Group Assignment} \label{sec:order_1}


In Eqn.~(\ref{eqn:gkmeans_obj}), 
each point is approximated by the summation of multiple codewords, 
and each codeword is chosen from a different dictionary.
We first take each dictionary as a group of clusters on 
the residual.
For the $c_1$-th dictionary, the residual is defined as
\begin{align}
\mathbf{y} = \mathbf{x} - 
\sum_{c \ne c_1}{\mathbf{d}^{c}_{k_c}}.
\label{eqn:residual}
\end{align}
The assignments $\{k_c, c\ne c_1\}$ are
from the previous iteration.
We can also assert that
the quantization error between $\mathbf{y}$
and any codeword $\mathbf{d}_{k_{c_1}}^{c_1} \in \mathcal{D}^{c_1}$
equals the distortion error between $\mathbf{x}$
and the combination of the codewords from all the dictionaries, i.e. $\|\mathbf{y} - \mathbf{d}^{c_1}_{k_{c_1}}\|^2_2 = \|\mathbf{x} - \sum_{c}\mathbf{d}_{k_{c}}^{c}\|^2_2$.
Thus, the index $k_{c_{1}}$ can naturally be chosen by
\begin{align}
k_{c_1}  = \arg\min_{k_{c_1}}\frac{1}{2}\|\mathbf{y} - \mathbf{d}_{k_{c_{1}}}^{c_1}\|_2^2.
\label{eqn:assignment_update}
\end{align}

Since the assignment $k_{c_{1}}$ depends on the assignment $k_{c}$ where $c \ne c_1$, 
we propose to  iteratively compute the assignments over 
all groups until
the assignments do not change. 
If the number of iterations needed to scan all the groups for one point is $S$, it requires $O(SKCP)$
multiplication.
To reduce the computation cost, we substitute Eqn.~(\ref{eqn:residual})
into Eqn.~(\ref{eqn:assignment_update}), and have
\begin{align}
\frac{1}{2}
\|\mathbf{y} - \mathbf{d}_{k_{c_1}}^{c}\|^2 
& =  
-\mathbf{x}^{T} \mathbf{d}_{k_{c_1}}^{c} + 
\sum_{c\ne c_1} {\mathbf{d}_{k_{c_1}}^{c_1}}^{T} 
\mathbf{d}_{k_c}^{c} + \frac{1}{2} {\mathbf{d}_{k_{c_{1}}}^{c_1}}^{T} \mathbf{d}_{k_{c_{1}}}^{c_1} 
+ \text{const}
\\
& = S_{k_{c_1}}^{c_1} + 
\sum_{c} T^{c_1, c}_{k_{c_1}, k_c} 
+ \text{const}
\label{eqn:expanded_assignment}
\end{align}
where $\text{const}$ is a constant to $k_{c_1}$ and 
\begin{align}
S_{k_{c_1}}^{c_1}  & = -\mathbf{x}^T \mathbf{d}_{k_{c_1}}^{c_1} \label{eqn:negative_inner_product} \\
T_{k_{c_1}, k_{c_2}}^{c_1, c_2} & = 
\begin{cases}
{\mathbf{d}^{c_1}_{k_{c_1}}}^{T} \mathbf{d}^{c_2}_{c_{k_2}} & c_1 \ne c_2 \\
\frac{1}{2}{\mathbf{d}^{c_1}_{k_{c_1}}}^{T} \mathbf{d}^{c_2}_{k_{c_2}} & c_1 = c_2,  k_{c_1} = k_{c_2} \\
0 & \text{otherwise}
\end{cases}
\label{eqn:T}.
\end{align}
The second item in Eqn.~(\ref{eqn:expanded_assignment})
is independent of the point $\mathbf{x}$, and 
thus we can pre-compute a lookup table $\{T_{k_{c_1}, k_{c_2}}^{c_1, c_2}\}$ before scanning all the points.
For each point, we compute $\{S_{k_{c_1}}^{c_1}\}$
before scanning all the group centers.
Then, the computation of Eqn.~(\ref{eqn:assignment_update}) only requires $O(C)$ addition
by Eqn.~(\ref{eqn:expanded_assignment}), 
and  the complexity of multiplication is reduced to $O(KCP)$  from $O(SKCP)$.
Since each dictionary is referred to as one group of clusters, we call this scheme \textit{Order-1 Group Assignment}
(shorted as $\text{O}_1\text{GA}$).


\begin{algorithm}[t]
\caption{Assignment Step in the current iteration}
\label{alg:assignment}
\begin{algorithmic}[1]
    \REQUIRE
        $\mathcal{X} = \{\mathbf{x}_i\}_{i = 1}^{N}$, $\{\mathbf{d}_{k}^{c}\}_{c = 1, k = 1}^{C, K}$, 
        assignments $\{k_{c}\}_{c = 1}^{C}$ for each $\mathbf{x} \in \mathcal{X}$ in the previous iteration.
    \ENSURE
        $\{k_{c}\}_{c = 1}^{C}$ for each $\mathbf{x} \in \mathcal{X}$ in the current iteration
    \STATE
    	Compute $\{T_{k_{c_1}, k_{c_2}}^{c_1, c_2}\}$ according to Eqn.~\ref{eqn:T}
    \FOR{$\mathbf{x} \in \mathcal{X}$}
    	\STATE
    		Compute $\{S_{k_{c_1}}^{c_1}\}_{c_1 = 1, k_{c_1} = 1}^{C, K}$ according to Eqn.~\ref{eqn:negative_inner_product}
    	\WHILE {! converged} \label{line:converged}
    		\STATE
    			Compute $k_{c_1},  c_1 \in \{1, \cdots, C\}$ by Sec.~\ref{sec:order_1} \\
    			or \\
    			Compute $(k_{c_1}, k_{c_2}), k_{c_2} = (k_{c_1} + 1) \% C,  c_1 \in \{1, \cdots, C\}$ by Sec.~\ref{sec:order_2}
        \ENDWHILE
    \ENDFOR
\end{algorithmic}
\end{algorithm}

\subsubsection{Order-$2$ Group Assignment} \label{sec:order_2}

Furthermore, we propose the scheme \textit{Order-2 Group Assignment} (shorted as $\text{O}_{2}\text{GA}$), where
any two dictionaries can be taken as a group of clusters. 

For the $c_1$-th and $c_2$-th dictionaries, we obtain the residual similarly as Eqn.~(\ref{eqn:residual})
by 
\begin{align}
\mathbf{y} = \mathbf{x} - \sum_{c\ne c_1, c\ne c_2}\mathbf{d}_{k_c}^{c}.
\label{eqn:assignment_group2}
\end{align}
Each pair of $(k_{c_1}, k_{c_2})$ can construct a cluster center $\mathbf{d}_{k_{c_1}}^{c_1} + \mathbf{d}_{k_{c_2}}^{c_2}$. 
Thus, we compute the assignment as
\begin{align}
(k_{c_1}, k_{c_2}) & = \arg\min_{k_{c_1}, k_{c_2}} \frac{1}{2} \|\mathbf{y} - \mathbf{d}_{k_{c_1}}^{c_1} - \mathbf{d}_{k_{c_2}}^{c_2}\|_2^2 \\
& = \arg \min_{k_{c_1}, k_{c_2}}(S_{k_{c_1}}^{c_1} + \sum_{c \ne c_2}T_{k_{c_1}, k_{c}}^{c_1, c}) + 
(S_{k_{c_2}}^{c_2} + \sum_{c \ne c_1}T_{k_{c_2}, k_{c}}^{c_2, c}) + 
T_{k_{c_1}, k_{c_2}}^{c_1, c_2},
\label{eqn:assignment_2}
\end{align}
where $S_{k_{c_1}}^{c_1}$ and $T_{k_{c_1}, k_{c_2}}^{c_1, c_2}$
are defined in Eqn.~(\ref{eqn:negative_inner_product}) and Eqn.~(\ref{eqn:T}), respectively.
To compute (\ref{eqn:assignment_2}), we need only $O(K^2)$ addition rather 
than multiplication. The formulation is general for any $c_1 \ne c_2$.
To reduce the time cost, we only apply it on two consecutive dictionaries.


\noindent\textbf{Discussion. }
The assignment algorithm is illustrated in Alg.~\ref{alg:assignment}
for $\text{O}_1\text{GA}$ and $\text{O}_2\text{GA}$. 
Note that the iteration in Line~\ref{line:converged}
is important because the assignments $ \{k_{c}, c \ne c_1\} $ or $\{k_c, c \ne c_1, c\ne c_2\}$ may change after all the 
dictionaries are scanned once,
and the assignment $k_{c_1}$ or $(k_{c_1}, k_{c_2})$ can be re-computed to further reduce the distortion error.

The scheme $\text{O}_{2}\text{GA}$ can be straightforwardly extended to $\text{O}_{n}\text{GA}$. 
If $n$ equals $C$, the global optimal assignment can be obtained. 
However, 
the time cost of all addition in 
 Eqn.~(\ref{eqn:assignment_2}) is $O(K^n)$  and is
exponential with $n$. 
Thus, we set $n = 1, 2$ experimentally.
Besides, the complexity is linear with the number of dictionaries $C$, 
and is much lower than the exponential complexity of \cite{WangWSXSL14}. 
Intuitively, the distortion induced by $\text{O}_{2}\text{GA}$
should be lower than or equal to that by $\text{O}_{1}\text{GA}$. 
However, this assertion cannot be guaranteed.
One reason is that in the iterative optimization, the dictionaries are also optimized 
and become quite different even after one update step. Given different dictionaries, we cannot 
assert the superiority of $\text{O}_{2}\text{GA}$ over $\text{O}_{1}\text{GA}$. However, in
most cases, the superiority is demonstrated experimentally. 

\subsection{Update Step} \label{sec:update}
The objective function in Eqn.~(\ref{eqn:gkmeans_obj})
is quadratic w.r.t. the multiple dictionaries. Thus, the dictionaries can be updated 
by setting the derivative w.r.t. the dictionaries as 0. In the following, we
derive the equivalent results from the mean of the centers, which is much similar with the 
traditional $K$-means.

Given the assignment $\{k_c\}$ for the $i$-th point, we first introduce an indicator function
\begin{align}
r_{i, k}^{c} = 
\begin{cases}
1, & k_{c} = k \text{  for } \mathbf{x}_i \\
0, & \text{otherwise.}
\end{cases}
\end{align}
It represents whether the assignment of the $i$-th point on the $c$-th dictionary is $k$.
Then, the residual in Eqn.~(\ref{eqn:residual}) can be written for the $i$-th point 
as 
\begin{align}
\mathbf{y}_i = \mathbf{x}_i - \sum_{c\ne c_1, k}r_{i, k}^{c}\mathbf{d}_{k}^{c}.
\label{eqn:residual_i}
\end{align}
Next, we update the center $\mathbf{d}_{k_1}^{c_1}$ by the mean of the residuals within the cluster, i.e.
\begin{align}
\mathbf{d}_{k_1}^{c_1} = \frac{\sum_{i} r_{i, k_1}^{c_1} \mathbf{y}_i  }{\sum_{i}r_{i, k_1}^{c_1}}, 
\text{if } \sum_{i}r_{i, k_1}^{c_1} \ne 0.
\label{eqn:update}
\end{align}
Substituting Eqn.~(\ref{eqn:residual_i}) into Eqn.~(\ref{eqn:update}) and simplifying the equation, we have
\begin{align}
\sum_{i}r_{i, k_1}^{c_1} \mathbf{x}_i = \sum_{k, c}\left(\sum_{i}r_{i, k_1}^{c_1} r_{i, k}^{c}\right) \mathbf{d}_k^c.
\label{eqn:one_ck}
\end{align}
Since Eqn.~(\ref{eqn:one_ck}) holds for any $c_1 \in \{1, \cdots, C\}$ and $k_1 \in \{1, \cdots, K\}$, 
the matrix form is $\mathbf{W} = \mathbf{D} \mathbf{Z}$, where
\begin{align}
\mathbf{W} & = 
\begin{bmatrix}
\mathbf{w}_{1}^{1} & \cdots & \mathbf{w}_{K}^{1} & \cdots & \mathbf{w}^{C}_{1} & \cdots & \mathbf{w}_{K}^{C}
\end{bmatrix} \\
\mathbf{w}_{k_1}^{c_1} & = \sum_{i}r_{i, k_1}^{c_1} \mathbf{x}_i, k_1 \in \{1, \cdots, K\}, c_1 \in \{1, \cdots, C\} \\
\mathbf{D} & = 
\begin{bmatrix}
\mathbf{d}_{1}^{1} & \cdots & \mathbf{d}_{K}^{1} & \cdots & \mathbf{d}^{C}_{1} & \cdots & \mathbf{d}^{C}_{K}
\end{bmatrix}.
\end{align}
The element of $\mathbf{Z} \in \mathbb{R}^{KC \times KC}$ in the $((c_1 - 1)K + k_1)$-th row and the $((c_2 - 1)K + k_2)$-th column 
is 
\begin{align}
Z_{k_1, k_2}^{c_1, c_2} = \sum_{i}r_{i, k_1}^{c_1} r_{i, k_2}^{c_2}, k_1, k_2 \in \{1, \cdots, K\}, c_1, c_2 \in \{1, \cdots, C\},
\end{align}
which can be interpreted as the number of points whose assignments on the $c_1$-th dictionary and on the 
$c_2$-th dictionary are $k_1$ and $k_2$, respectively.
Then, we can solve the dictionaries $\mathbf{D}$ by the matrix (pseudo)inversion.

\subsection{Initialization}\label{sec:initialization}
Since the problem of minimizing Eqn.~(\ref{eqn:gkmeans_obj}) is non-convex, 
different initializations fall 
into different local minima,
and thus the initialization is quite important.

\subsubsection{Initialization of  Dictionaries} \label{sec:init_dictionary}

One direct method is to randomly sample the points from the dataset to construct
the dictionaries as in \cite{WangWSXSL14}. 
Empirically, we find this scheme works well for cases with a small number
of dictionaries (e.g. $C = 2$ as in~\cite{WangWSXSL14}), 
but the performance degrades for a larger $C$.
In this work, we propose two initialization schemes. The
first is based on the traditional $K$-means and the second is a hierarchical scheme based 
on~\cite{NorouziF13, WangWSXSL14}.

\noindent\textbf{$K$-Means-based initialization} is to run the traditional $K$-means algorithm
on the residual repeatedly. Specifically, 
we first set the residual by the original dataset, i.e. $\mathbf{y} = \mathbf{x}, \forall\mathbf{x} \in \mathcal{X}$, and the index of the to-be-initialized dictionary as $c = 1$.
Then, the $K$-means is performed on $\{\mathbf{y}\}$
to obtain the $K$ cluster centers as the first dictionary $\{\mathbf{d}^{c}_1, \cdots, \mathbf{d}^{c}_{K}\}$. 
After this, we update the residual of each point by $\mathbf{y} \leftarrow \mathbf{y} - \mathbf{d}_{k_c}^{c}$,
where $k_c$ is the assignment of $\mathbf{y}$, 
and $c \leftarrow c + 1$. 
The $K$-means algorithm is repeatedly run to obtain the second dictionary. 
By alternately updating the residual and performing the $K$-means on the residual, we can 
get all the dictionaries as an initialization.

\noindent\textbf{Hierarchical Initialization} 
can ensure that
in theory
the approach performs not worse than the vector partitioning approaches (e.g.~\cite{NorouziF13})
under the same code length.
The basic idea is to solve $\log_{2}{C}$\footnote{We assume $C$ is a power of $2$. Meanwhile, the dimension $P$ is assumed to be divisible by $C$ in the following. 
This algorithm can easily adapt to general cases.}
 subproblems where different constraints are applied 
on the multiple dictionaries. 
Let $\mathbf{D}^{c} = \begin{bmatrix}
\mathbf{d}^{c}_{1} & \cdots & \mathbf{d}^{c}_K
\end{bmatrix}$ and $\mathbf{b}^{c} \in \{0, 1\}^{K}, \|\mathbf{b}^{c}\|_1 = 1$. 
The index of $1$ in $\mathbf{b}^{c}$ represents which codeword is selected.
We can rewrite the objective of Eqn.~(\ref{eqn:gkmeans_obj})
as 
\begin{align}
\sum_{\mathbf{x}} 
\min_{\mathbf{b}^{c}, \|\mathbf{b}^{c}\|_1 = 1}
\left\|\mathbf{x} - \begin{bmatrix}
\mathbf{D}^{1} & \cdots & \mathbf{D}^{C}
\end{bmatrix} 
\begin{bmatrix}
\mathbf{b}^{1} \\
\vdots \\
\mathbf{b}^{C}
\end{bmatrix}
\right\|_2^2.
\label{eqn:gkmeans_obj_matrix}
\end{align}

Before introducing the approach  in a general case, we take $C = 4$
as an example. 
First, we minimize Eqn.~(\ref{eqn:gkmeans_obj_matrix})
with the dictionaries $\mathbf{D} = \begin{bmatrix}
\mathbf{D}^{1} & \mathbf{D}^{2} & \mathbf{D}^{3} & \mathbf{D}^{4}
\end{bmatrix}$ constrained to be
\begin{align}
\mathbf{D} = 
\mathbf{R}_1
\begin{bmatrix}
\mathbf{D}_1^{1, 1}& \mathbf{0} & \mathbf{0} & \mathbf{0} \\
\mathbf{0} & \mathbf{D}_1^{2, 1} & \mathbf{0} & \mathbf{0} \\
\mathbf{0} & \mathbf{0} & \mathbf{D}_1^{3, 1} & \mathbf{0} \\
\mathbf{0} & \mathbf{0} & \mathbf{0} & \mathbf{D}_1^{4, 1} 
\end{bmatrix}, \mathbf{R}_1^{T} \mathbf{R}_1 = \mathbf{I},
\label{eqn:R_ckmeans}
\end{align}
where $\mathbf{R}_1$ is a rotation matrix and $\mathbf{D}_{1}^{m, 1}, m\in \{1, 2, 3, 4\}$
is a real matrix of size $P / 4 \times K$.
The notation $\mathbf{D}_{s}^{u, v}$ denotes the block in the $u$-th row and $((u - 1)2^{s - 1} + v)$-th column
of $\mathbf{D}$ in the $s$-th subproblem.
This subproblem is studied in~\cite{NorouziF13}, and 
can be solved by alternating optimizations with regard to $\mathbf{R}_1$, $\{\mathbf{D}_1^{m, 1}\}$
and $\{\mathbf{b}^{c}\}$.
We initialize $\mathbf{R}_1$ by an identity matrix and $\{\mathbf{D}_{1}^{m, 1}\}$
by randomly choosing the data points on the corresponding subvector.
The optimal solution is used to initialize the second subproblem where the objective function remains the same
but the constraint is relaxed to be
\begin{align}
\mathbf{D} = 
\mathbf{R}_2
\begin{bmatrix}
\mathbf{D}_2^{1, 1}& \mathbf{D}_{2}^{1, 2} & \mathbf{0} & \mathbf{0} \\
\mathbf{0} & \mathbf{0} & \mathbf{D}_2^{2, 1} & \mathbf{D}_2^{2, 2}
\end{bmatrix}, {\mathbf{R}_2}^{T} \mathbf{R}_2 = \mathbf{I}.
\end{align}
The initialization of the second subproblem is as follows
\begin{align}
\mathbf{R}_{2} = \mathbf{R}_{1}^{*},
\mathbf{D}_{2}^{1, 1} = \begin{bmatrix}
{\mathbf{D}_{1}^{1, 1}}^{*} \\
\mathbf{0}
\end{bmatrix}, 
\mathbf{D}_{2}^{1, 2} = \begin{bmatrix}
\mathbf{0} \\
{\mathbf{D}_{1}^{2, 1}}^{*}
\end{bmatrix},
{\mathbf{D}_{2}^{2, 1}} = \begin{bmatrix}
{\mathbf{D}_{1}^{3, 1}}^{*} \\
\mathbf{0}
\end{bmatrix},
\mathbf{D}_{2}^{2, 2} = \begin{bmatrix}
\mathbf{0} \\
{\mathbf{D}_{1}^{4, 1}}^{*}
\end{bmatrix},
\end{align}
where the asterisk $^*$ denotes the optimal solution.
This subproblem is studied in~\cite{WangWSXSL14}, 
and it is verified that the distortion can be lower than ck-means~\cite{NorouziF13}.
We solve the subproblem in a similar manner with~\cite{WangWSXSL14},
except that the assignment step is replaced by that in Sec.~\ref{sec:assignment}.
Finally $\mathbf{D}$ is initialized by
\begin{align}
\mathbf{D}^{1} = \mathbf{R}_2^{*}\begin{bmatrix}
{\mathbf{D}^{1, 1}_{2}}^{*} \\
\mathbf{0}
\end{bmatrix},
\mathbf{D}^{2} = \mathbf{R}_{2}^{*}\begin{bmatrix}
{\mathbf{D}_{2}^{1, 2}}^{*} \\
\mathbf{0}
\end{bmatrix}, 
\mathbf{D}^{3} = \mathbf{R}_{2}^{*} \begin{bmatrix}
\mathbf{0} \\
{\mathbf{D}_{2}^{2, 1}}^{*} 
\end{bmatrix},
\mathbf{D}^{4} = \mathbf{R}_{2}^{*} \begin{bmatrix}
{\mathbf{D}_{2}^{2, 2}}^{*} \\
\mathbf{0}
\end{bmatrix}.
\end{align}

In summary, each subproblem enforces different levels of restrictions on the multiple dictionaries. 
In the first subproblem, the restriction is most severe, and most of the elements are constrained to be
$0$. Then it is relaxed gradually in the subsequent subproblems.
Generally, the $s$-th ($s \in \{1, \cdots, \log_{2}{C}\}$) subproblem
is constrained by ${\mathbf{R}_{s}}^{T} \mathbf{R}_{s} = \mathbf{I}$ and
\begin{align}
\mathbf{D} = 
\mathbf{R}_{s}
\begin{bmatrix}
\mathbf{D}_{s}^{1, 1} & \cdots & \mathbf{D}_{s}^{1, 2^{s -1}} &  &  & \\
 &  & & \ddots & & & \\
 & & & & & & \mathbf{D}_{s}^{C/2^{s - 1}, 1} & \cdots & \mathbf{D}_{s}^{C/2^{s - 1}, 2^{s - 1}}
\end{bmatrix}.
\end{align}
The initialization of the $(s + 1)$-th ($s \le \log_{2}{C} - 1$)
subproblem from the optimal solution of the $s$-th subproblem is $\mathbf{R}_{s + 1} = {\mathbf{R}_s}^{*}$
and
\begin{align}
\mathbf{D}_{s + 1}^{u, v} & = \begin{bmatrix}
{\mathbf{D}_{s}^{2u - 1, v}}^{*} \\
\mathbf{0}
\end{bmatrix},  ~~~~~\text{if }  u \in \{1, \cdots, {C}/{2^s}\}, v \in \{1, \cdots, 2^{s - 1}\}; \\
\mathbf{D}_{s + 1}^{u, v} & = \begin{bmatrix}
\mathbf{0} \\
{\mathbf{D}_{s}^{2u, v - 2^{s - 1}}}^{*}
\end{bmatrix}, \text{if  } u \in \{1, \cdots, {C}/{2^s}\}, v \in \{2^{s - 1} + 1, \cdots, 2^{s}\}.
\end{align}
The final initialization for Eqn.~(\ref{eqn:gkmeans_obj}) is
\begin{align*}
\mathbf{D}^{v}  = 
\mathbf{R}^{*}_{\log_2 C}
\begin{bmatrix}
{\mathbf{D}_{\log_2 C}^{1, v}}^{*} \\
\mathbf{0}
\end{bmatrix}, 1 \le v \le C / 2;
\mathbf{D}^{v}  = 
\mathbf{R}_{\log_2 C}^{*}
\begin{bmatrix}
\mathbf{0} \\
{\mathbf{D}_{\log_2 C}^{2, v - C / 2}}^{*}
\end{bmatrix},  C/2 + 1 \le v \le C
\end{align*}

\subsubsection{Initialization of assignments} \label{sec:init_assignment}
In Sec.~\ref{sec:assignment}, the assignment is based on the residual defined in Eqn.~(\ref{eqn:residual})
and Eqn.~(\ref{eqn:assignment_group2}) for $\text{O}_{1}\text{GA}$ and $\text{O}_{2}\text{GA}$, respectively.
To initialize the assignment, we only use the initialized $k_c$ to compute the residual. 
Taking $\text{O}_{1}\text{GA}$ as an example, we compute the residual by $\mathbf{y} = \mathbf{x} - \sum_{c = 1}^{c_1 - 1}\mathbf{d}_{k_c}^{c}$ for the $c_1$-th dictionary.
After iterating $c_1$ from $1$ to $C$, we
refine the assignment by Alg.~\ref{alg:assignment}. 
This is also applied to encode new data points after obtaining the dictionaries. That is, we initialize
the assignments first and then refine them by Alg.~\ref{alg:assignment}.
Similar ideas can be applied to initialize the assignments for $\text{O}_{2}\text{GA}$.

\section{Experiments}
\subsection{Settings}
We conduct the experiments on three widely-used high-dimensional datasets:
SIFT1M~\cite{JegouDS11},
GIST1M~\cite{JegouDS11}, and MNIST~\cite{LeCunBBH98}.
SIFT1M has $10^5$ training features, $10^4$ query features, and $10^6$ database features. 
Each feature is a $128$-dimensional SIFT descriptor.
GIST1M has $5\times 10^5$ training features, $10^3$ query features and $10^6$ database features with each being a $960$-dimensional GIST feature. MNIST contains $60,000$ database images (also used as the training set) and $10,000$ query images. 
Each image has $28\times 28$ pixels, and we vectorize it as a $784$-dimensional 
feature vector. 

The accuracy is measured by the relative distortion,
which 
is defined as the objective function of Eqn.~(\ref{eqn:gkmeans_obj}) with the optimized solutions divided 
by $\sum_{\mathbf{x} \in \mathcal{X}} \|\mathbf{x}\|_2^2$.
This indicator is important in data compression, approximate nearest neighbor (ANN) search~\cite{WangWSXSL14}, etc.
The accuracy is better with a lower distortion.
Due to the space limitation, we 
report the experiments on the application of ANN search in the supplementary material.

\begin{figure}[t]
\centering
\begin{tabular}{c@{~}c@{~}c}
32 bits & 64 bits & 128 bits \\
\includegraphics[width = \sizetraining\linewidth]{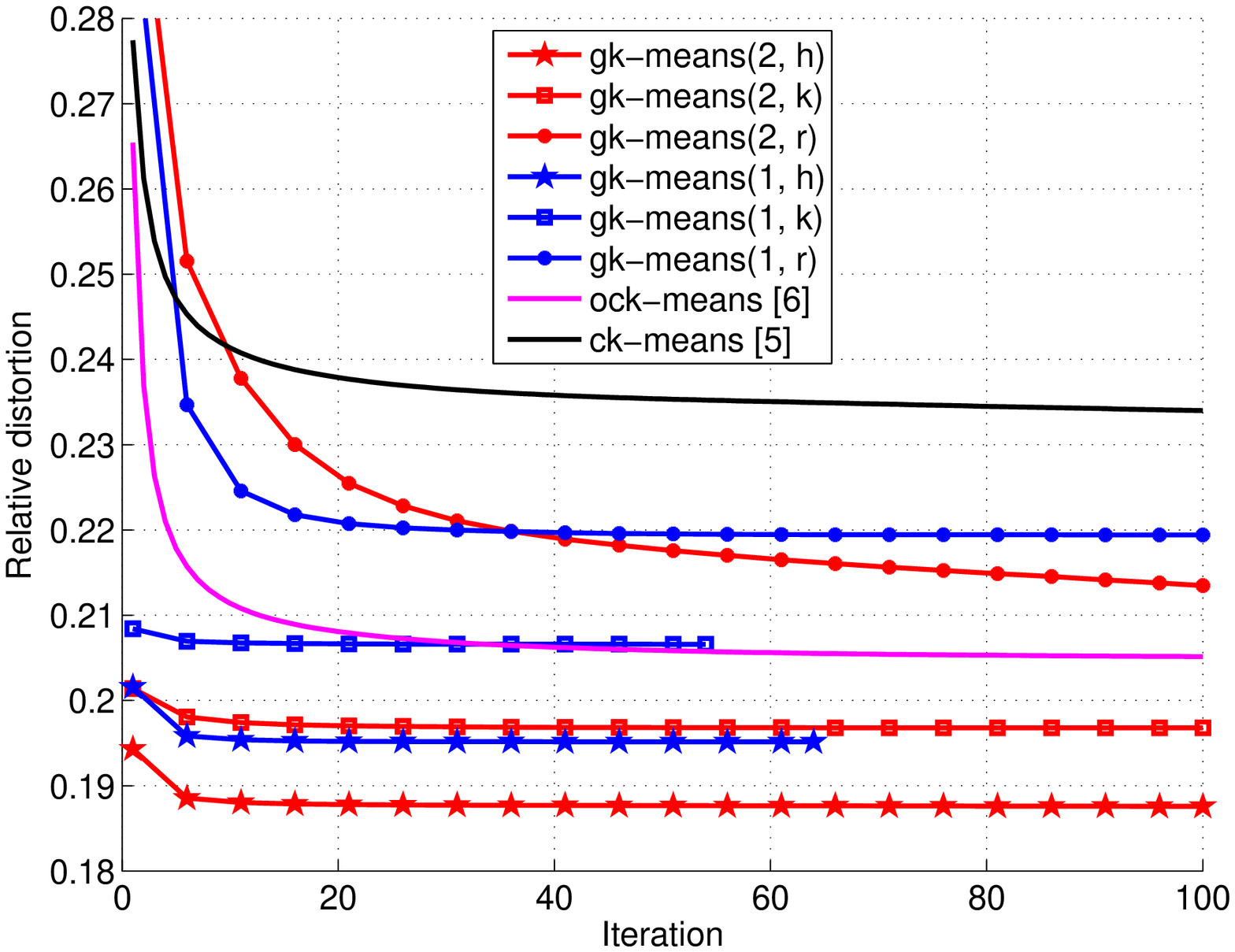} & 
\includegraphics[width = \sizetraining\linewidth]{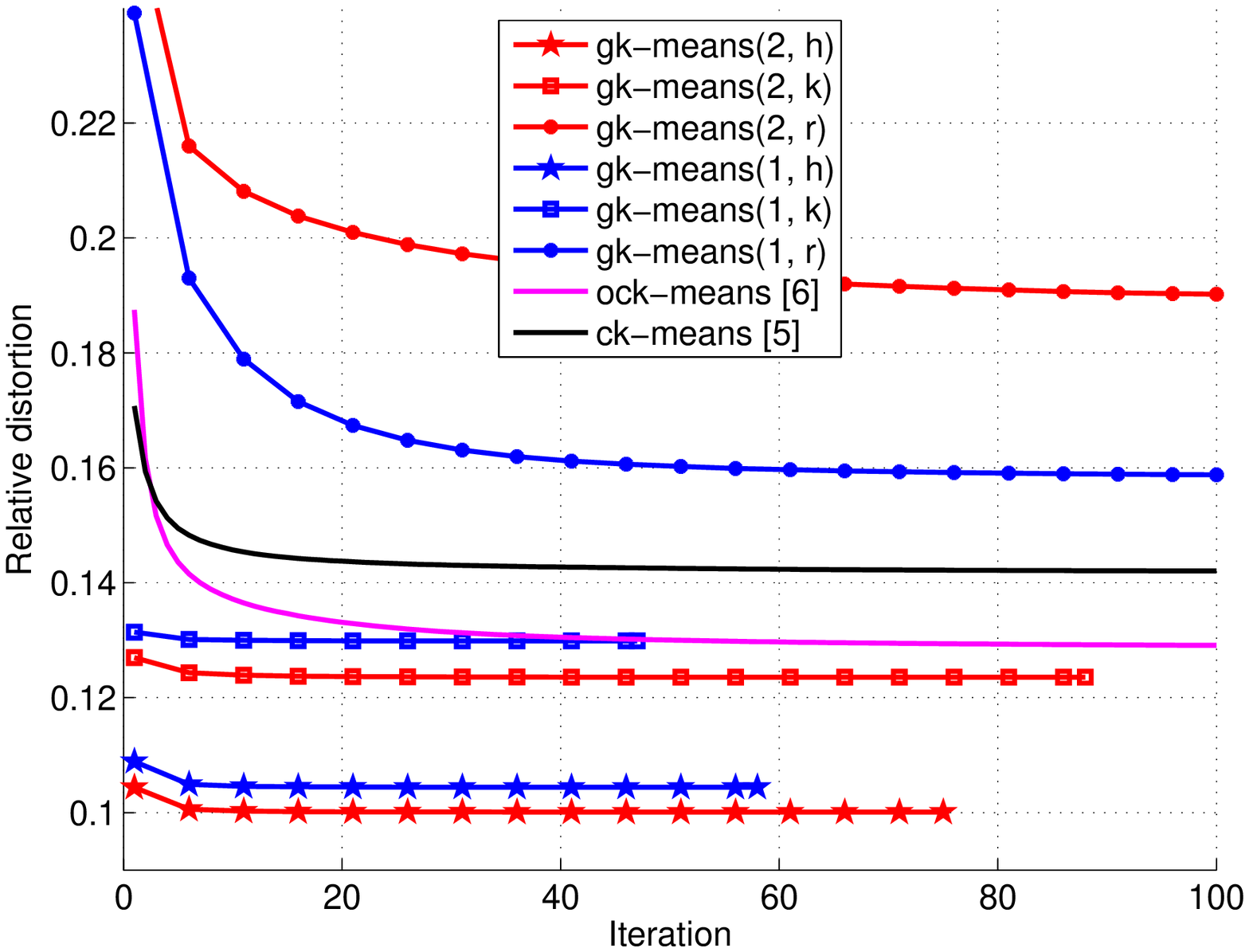} & 
\includegraphics[width = \sizetraining\linewidth]{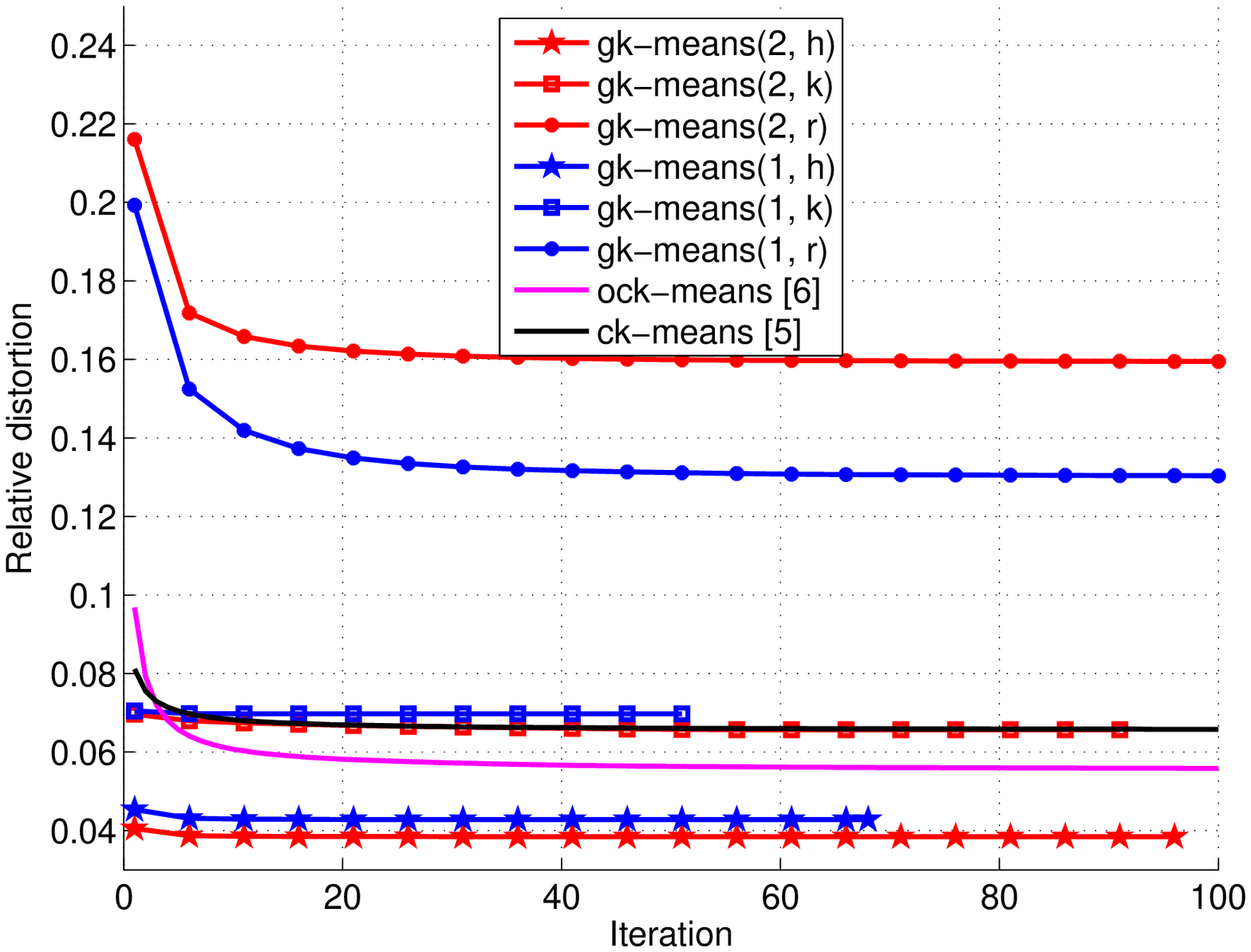} \\
(a) & (b) & (c) \\
\includegraphics[width = \sizetraining\linewidth]{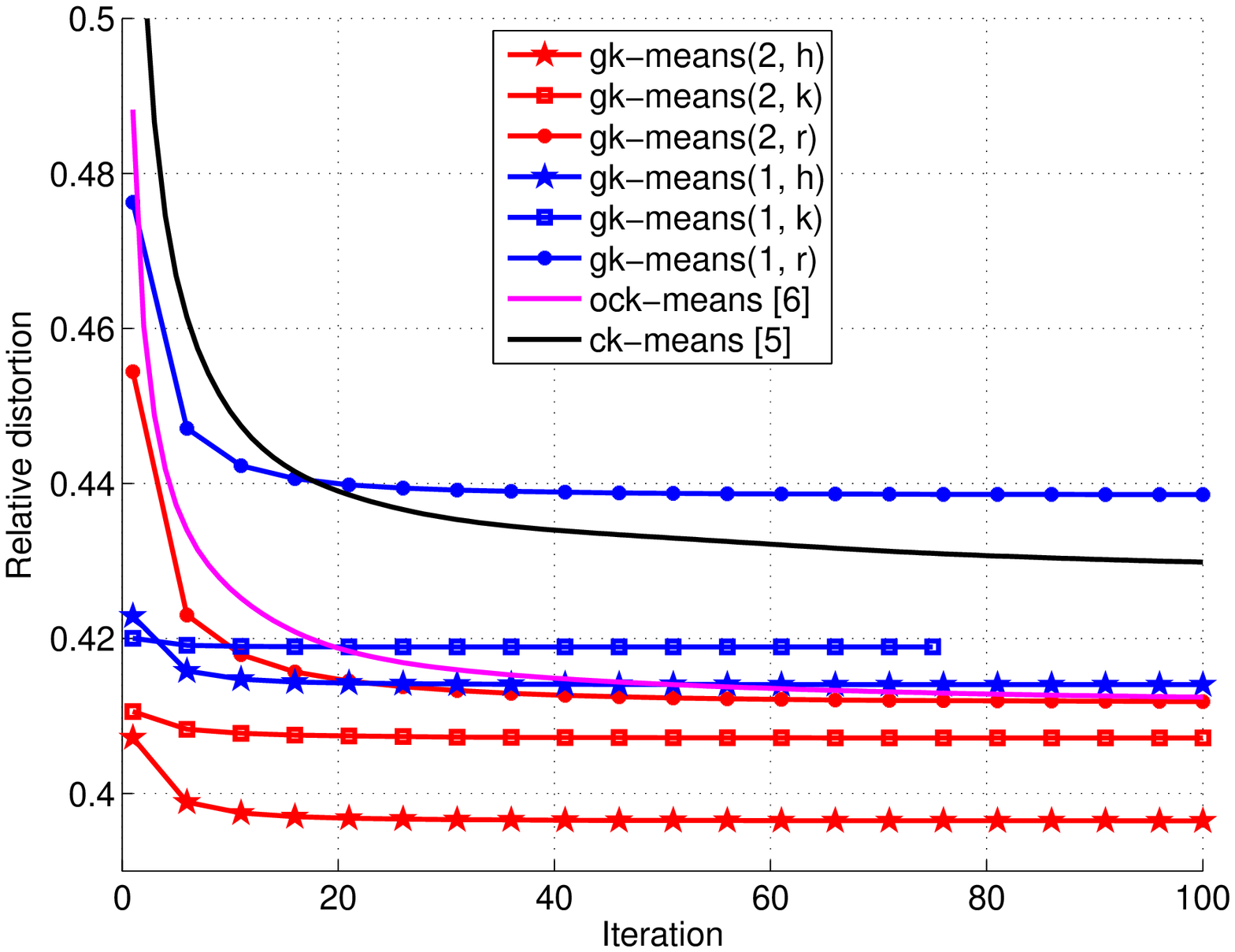} & 
\includegraphics[width = \sizetraining\linewidth]{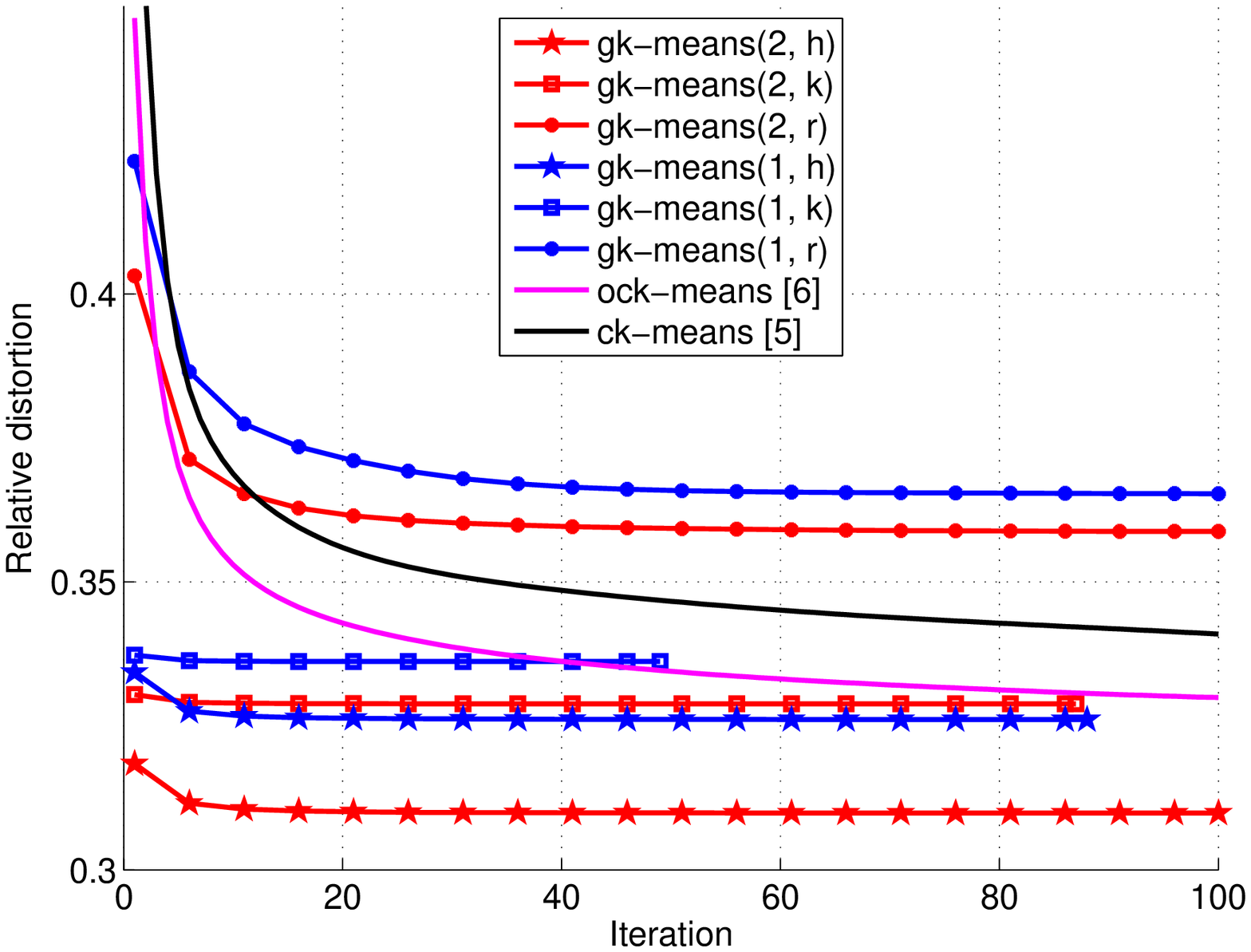} & 
\includegraphics[width = \sizetraining\linewidth]{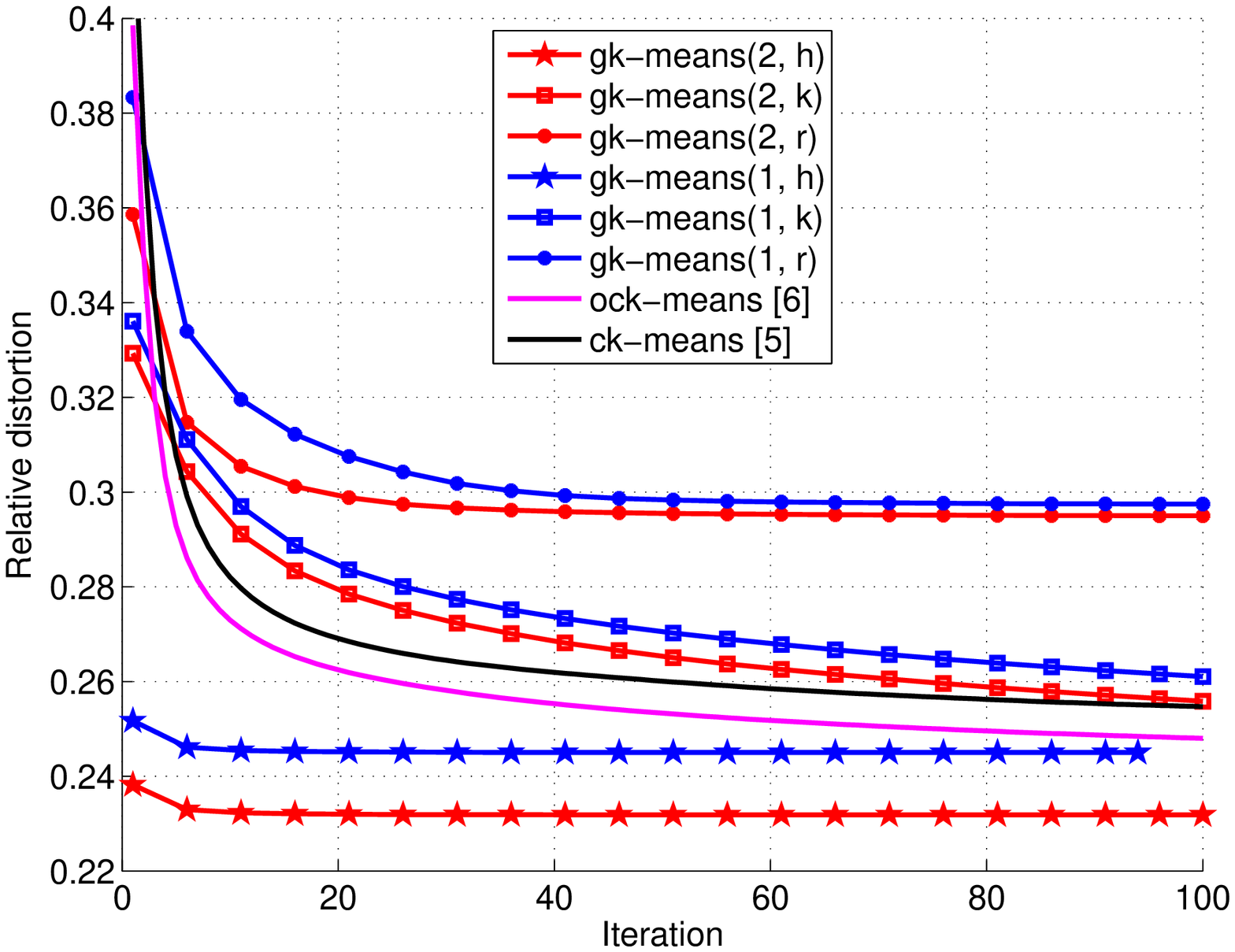} \\
(d) & (e) & (f) \\
\includegraphics[width = \sizetraining\linewidth]{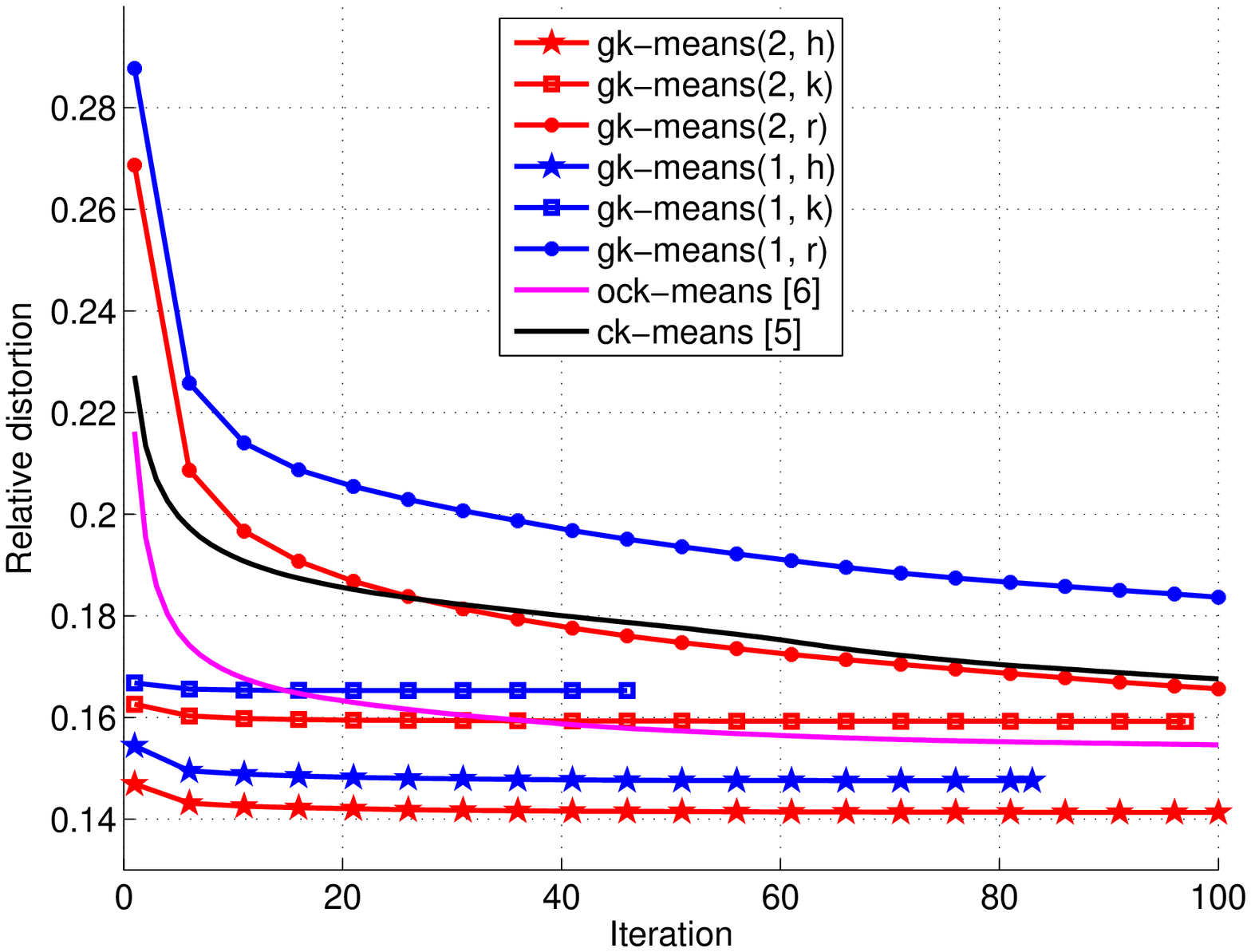} & 
\includegraphics[width = \sizetraining\linewidth]{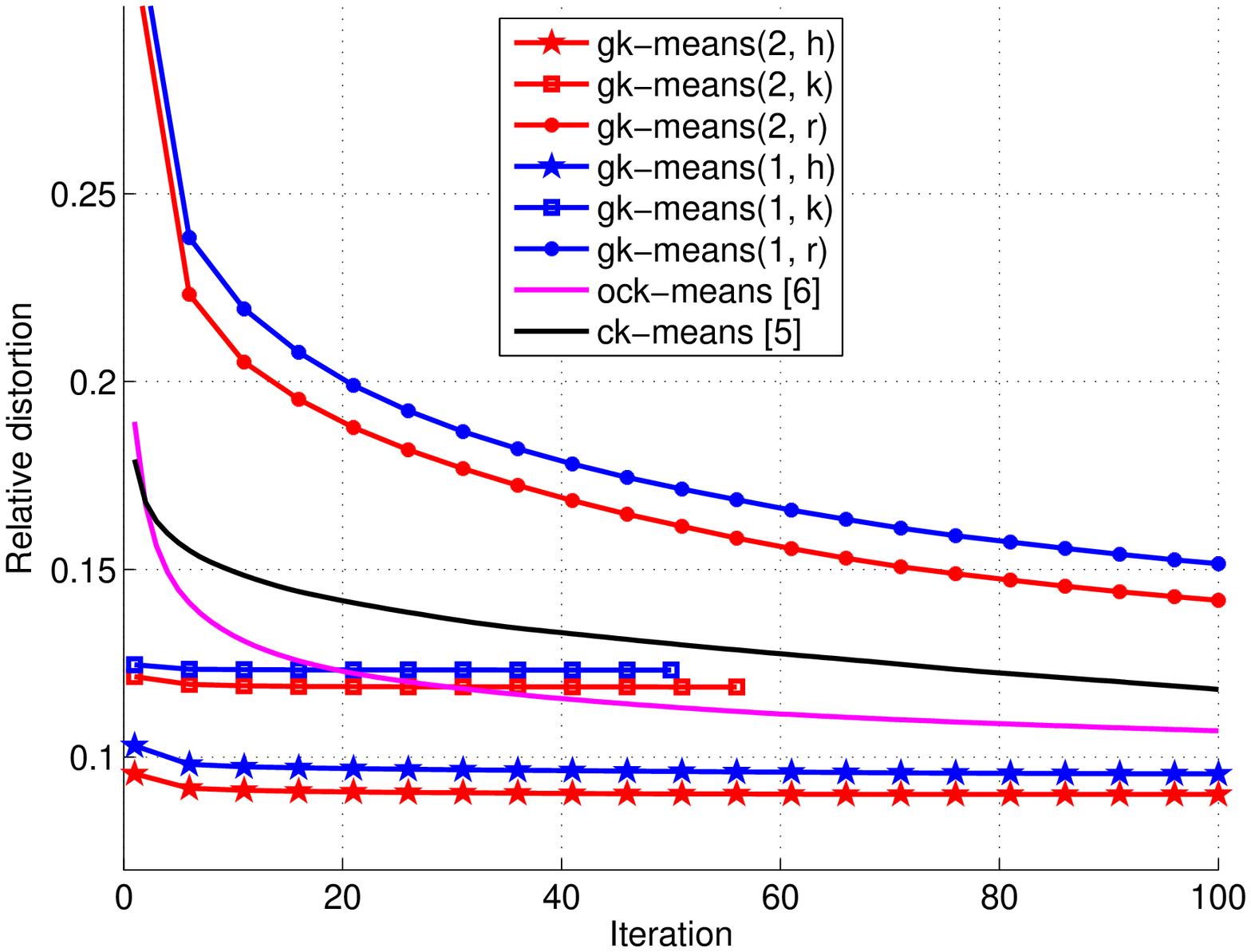} & 
\includegraphics[width = \sizetraining\linewidth]{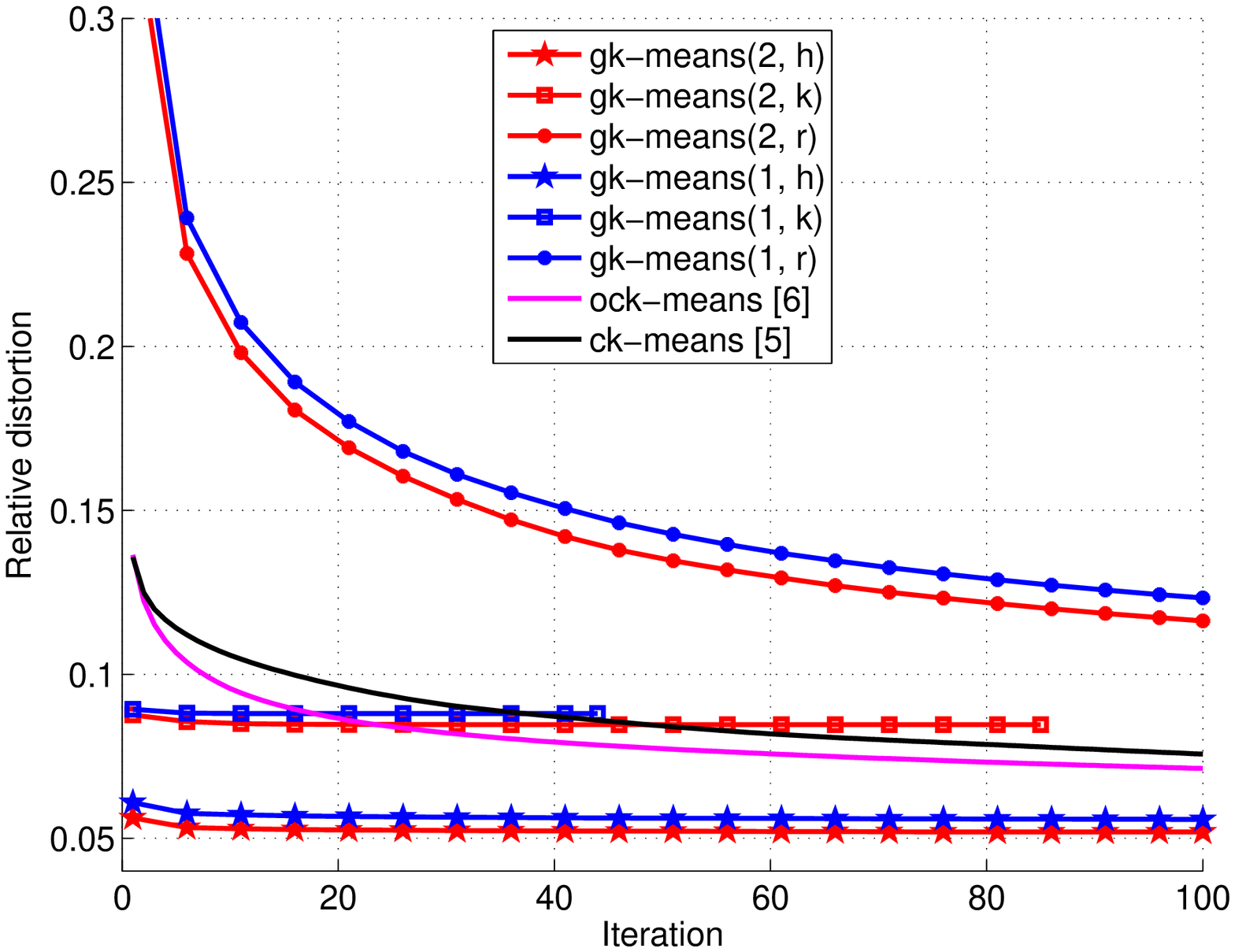} \\
(g) & (h) & (i)
\end{tabular}
\caption{Relative distortion on the training set 
with different numbers of iterations.
The first row corresponds to SIFT1M; the second to GIST1M; and 
the third to MNIST.}
\label{fig:relative_distortion_training}
\end{figure}

We compare our approach with ck-means~\cite{NorouziF13} and
optimized Cartesian $k$-means(ock-means)~\cite{WangWSXSL14}
 under the same code length.
Let $M_{\text{ck}}$ and $M_{\text{ock}}$ be 
the numbers of subvectors in ck-means and ock-means respectively,
and let $C_{\text{ock}}$ and $C_{\text{gk}}$
be the number of dictionaries on each subvector of ock-means and 
the number of dictionaries of gk-means respectively.
Then the code lengths of ck-means, ock-means, and gk-means
are
$M_{\text{ck}}\log_2(K)$, $M_{\text{ock}}C_{\text{ock}}\log_{2}(K)$
and $C_{\text{gk}}\log_{2}(K)$, respectively. 
We set $M_{\text{ck}} = M_{\text{ock}} C_{\text{ock}} = C_{\text{gk}}$
to make the code length identical for all the approaches. 
Following~\cite{NorouziF13, WangWSXSL14}, we set $K = 256$
to fit the index by one byte,
and $C_{\text{gk}} = 4, 8, 16$ to obtain the code lengths $32, 64, 128$, respectively.
As for the initialization of gk-means shown in Sec.~\ref{sec:init_dictionary},
$30$ iterations are consumed both in each $k$-means for the $k$-means-based initialization
and in each subproblem for the hierarchical initialization.
The number of iterations in all the approaches is at most $100$ 
or the iteration stops if it reaches convergence.
It is expected that the performance gains with a larger number of iterations. Here we just fix the 
maximum iteration number for the comparison purpose.
To minimize Eqn.~(\ref{eqn:gkmeans_obj}), a multiple candidate matching pursuit (MCMP) algorithm
is proposed on each subvector in \cite{WangWSXSL14}. 
Since MCMP is exponential with the number of dictionaries, we set 
$C_{\text{ock}} = 2$ by default as suggested in~\cite{WangWSXSL14}.
We also run MCMP on the full vector with other values of $C_{\text{ock}}$
and compare it with gk-means.

The term \textit{gk-means(a, b)} is used
to distinguish different
assignment approaches in Sec.~\ref{sec:assignment}
and different initialization schemes in Sec.~\ref{sec:initialization}:
a = 1, 2 to represent $\text{O}_{1}\text{GA}$ and $\text{O}_{2}\text{GA}$, respectively; 
b = r, k, h to represent the random initialization, the k-means-based initialization,
and the hierarchical initialization, respectively.

\subsection{Results}

The relative distortions on the  training sets
are illustrated in Fig.~\ref{fig:relative_distortion_training}
with different numbers of iterations.
In each iteration, we report the relative distortion after Line~\ref{line:assignment} in Alg.~\ref{alg:overall}.
Since ock-means and ck-means adopt a similar alternating optimization algorithm, we also collect 
the relative distortions before the end of each iteration.
Certain curves stop before $100$ iterations because of convergence.

\noindent\textbf{$\text{O}_{1}\text{GA}$ vs $\text{O}_{2}\text{GA}$.}
In most cases,  $\text{O}_{2}\text{GA}$
is better than $\text{O}_{1}\text{GA}$ except for certain cases, e.g.,  
the comparison of gk-means(1, r) and gk-means(2, r) in 
Fig.~\ref{fig:relative_distortion_training} (b).
This may be because in the iterative optimization, 
different assignment algorithms generate different assignments, and 
the dictionaries are optimized in different directions.
With different dictionaries, we cannot guarantee the superiority of $\text{O}_2\text{GA}$
over $\text{O}_1\text{GA}$.
In the cases where the $\text{O}_{2}\text{GA}$ is better, the improvement varies
among different settings. 
For example in Fig.~\ref{fig:relative_distortion_training} (a) and Fig.~\ref{fig:relative_distortion_training} (d), the improvement of gk-means(2, h) over gk-means(1, h) is much more significant,
while in Fig.~\ref{fig:relative_distortion_training} (b) and Fig.~\ref{fig:relative_distortion_training} (c), 
the difference is minor. 

\noindent\textbf{Comparison of initializations.}
Generally, the hierarchical initialization is the best; the second 
is the $k$-means-based, and the worst is the random initialization.
Although we cannot guarantee this in theory, 
this observation is true under all the settings in practice
in Fig.~\ref{fig:relative_distortion_training}.

\noindent\textbf{Comparison with ock-means and ck-means.}
From Fig.~\ref{fig:relative_distortion_training}, we can see that 
the random initialization
is almost always inferior, 
while the hierarchical initialization can always 
lead to a lower distortion than ock-means and ck-means.
This also implies that
the initialization is quite important to such a 
non-convex problem.

Similar observations can be found w.r.t. the relative distortion 
on the databases illustrated in Table~\ref{tbl:base_loss}.

Finally, we compare the gk-means and the MCMP~\cite{WangWSXSL14}
on the full vector ($M_{\text{ock}} = 1$). 
We evaluate the time cost on a Linux server 
with a CPU of 2660MHz and 48G memory.
The experiment is conducted on SIFT1M, 
and the program runs in $24$ threads to encode the $10^6$ database points.

The results are depicted in Table~\ref{fig:comp_mcmp}. 
Due to the high time cost, we cannot run MCMP with $8$ dictionaries. Thus, 
we estimate the time cost as follows. The dictionaries are 
trained by gk-means(2, h), and the time cost is collected on $240$
database points with the MCMP algorithm since we deploy $24$ threads. 
The result is $2589.66$ seconds and we multiply it by $10^6 / 240$
to estimate the time cost to encode the whole $10^6$ database points.
The results of gk-means are tested with the best hierarchical initialization.
From the results, we can see that in terms of time cost both gk-means(1, h)
and gk-means(2, h) scale well with the number of dictionaries while 
MCMP does not.
This is because the time cost of gk-means is linear to the number of dictionaries
while MCMP is exponential.
Meanwhile, the time cost of gk-means(1, h) is less than that of gk-means(2, h),
because the complexity of the former is $O(K)$ while that of the latter is $O(K^2)$.
In terms of the relative distortion, MCMP is slightly better
than gk-means(2, h) which is better than gk-means(1, h).

\begin{table}
\centering
\caption{Relative distortion ($\times 10^{-2}$) on the databases with different code lengths.}
\label{tbl:base_loss}
\begin{tabular}{c@{~~}cccccccc@{~~}c}
\toprule
\multicolumn{2}{c}{} &
\multicolumn{6}{c}{gk-means} & 
\multirow{2}{*}{ock-means } & 
\multirow{2}{*}{ck-means} \\ 
\cmidrule{3-8} 
\multicolumn{2}{c}{} & (2, h) & (2, k) & (2, r) & (1, h) & (1, k) & (1, r) & & \\ 
\midrule
\multirow{3}{*}{SIFT1M}  & 32  & $\mathbf{19.68}$  & 20.47  & 22.22  & 20.63  & 21.59  & 22.48  & 21.14  & $\textit{23.78}$ \\
 & 64  & $\mathbf{11.01}$  & 13.19  & $\textit{19.46}$  & 11.56  & 14.10  & 16.16  & 13.37  & 14.45 \\
 & 128  & $\mathbf{4.54}$  & 7.44  & $\textit{16.92}$  & 5.08  & 8.25  & 12.62  & 5.85  & 6.71 \\
\midrule
\multirow{3}{*}{GIST1M}  & 32  & $\mathbf{40.56}$  & 41.24  & 41.58  & 42.73  & 42.59  & $\textit{44.32}$  & 41.64  & 43.19 \\
 & 64  & $\mathbf{32.11}$  & 33.86  & 35.94  & 33.92  & 34.75  & $\textit{36.76}$  & 33.38  & 34.32 \\
 & 128  & $\mathbf{24.48}$  & 26.87  & 29.12  & 25.93  & 27.56  & $\textit{29.29}$  & 25.18  & 25.69 \\
\midrule
\multirow{3}{*}{MNIST}  & 32  & $\mathbf{14.25}$  & 16.52  & 17.62  & 15.29  & 18.18  & $\textit{22.43}$  & 15.53  & 16.76 \\
 & 64  & $\mathbf{9.44}$  & 14.12  & 16.48  & 10.36  & 14.88  & $\textit{18.38}$  & 10.83  & 11.81 \\
 & 128  & $\mathbf{5.91}$  & 11.63  & 14.28  & 6.55  & 12.37  & $\textit{16.11}$  & 7.28  & 7.57 \\
\bottomrule
\end{tabular}
\end{table}

\begin{table}
\centering
\caption{Encoding time and the relative distortion (R.d) on the database of SIFT1M.}
\label{fig:comp_mcmp}
\begin{tabular}{cccccccc}
\toprule
\multicolumn{2}{c}{} & \multicolumn{2}{c}{2} & \multicolumn{2}{c}{4} & \multicolumn{2}{c}{8} \\
\cmidrule(lr){3-4}  \cmidrule(lr){5-6} \cmidrule(lr){7-8}
\multicolumn{2}{c}{} & Time (s) & R.d. & Time  (s) & R.d. & Time (s) & R.d \\
\midrule
\multirow{2}{*}{gk-means} & 1 
 & 8.6 
 & 0.30
 & 20.3 
 & 0.21
 & 120.4 
 & 0.12
\\
 & 2 
 & 45.2 
 & 0.29
 & 110.3 
 & 0.20
 & 392.9 
 & 0.11
\\
\midrule
\multicolumn{2}{c}{MCMP} 
 & 12.2 
 & 0.29
 & 723.3 
 & 0.19
 & 10790250.0 
 & --
\\
\bottomrule
\end{tabular}
\end{table}


\section{Conclusion}
We proposed a simple yet effective algorithm, named as group $k$-means,
to effectively encode the data point.
With the desirable low distortion errors, this approach can represent high-dimensional data points
with less storage cost and facilitate data management.
Future work includes applying it to applications, e.g., multimedia retrieval with
inner product measurement and image compression.

\begin{appendices}

This appendix reports
the experimental results of approximate nearest
neighbor search, in which our 
group $K$-means (gk-means) is compared 
with other approaches.

Following~\cite{WangWSXSL14}, 
we compute the approximate distance
between the query $\mathbf{q}\in \mathbb{R}^{P}$
and the database point $\mathbf{x} \in \mathbf{R}^{P}$ 
encoded as $(k_1, \cdots, k_C)$ by
\begin{align}
\frac{1}{2}\left\|\mathbf{q} - \sum_{c} \mathbf{d}^{c}_{k_c} \right\|_2^2 = 
\frac{1}{2}\|\mathbf{q}\|_2^2 - 
\sum_{c}{\mathbf{q}^{T} \mathbf{d}^{c}_{k_c}} + 
\left\|\sum_{c} \mathbf{d}^{c}_{k_c} \right\|_2^2
\label{eqn:asy_distance}.
\end{align}
The first item is consistent and can be omitted during distance evaluation. 
The third item is independent of the query and 
can be pre-computed from the code $(k_1, \cdots, k_{C})$
and the dictionaries. No original data is required for the 
computation of the third item.
To compute the second item, we can evaluate all the 
inner products
$\mathbf{q}^{T} \mathbf{d}^{c}_{k}$
for $c\in \{1, \cdots, C\}$ and $k \in \{1, \cdots, K\}$. 
Then, each distance computation only involves $O(C) + 1$ addition, which
is comparable to ck-means~\cite{NorouziF13}.

For each query, we compare it with every database point by Eqn.~\ref{eqn:asy_distance},
and rank all the points by the approximate distance. 
We take recall as the  performance criterion to measure the proportion of the queries  
whose corresponding true nearest neighbors (by Euclidean distance)
fall in the top ranked points.

As studied in the paper, initialization is important for the minimization of the objective function
and we only report the results with the best hierarchical initialization scheme. 

\begin{figure}[t]
\centering
\begin{tabular}{ccc}
\includegraphics[width = 0.3\linewidth]{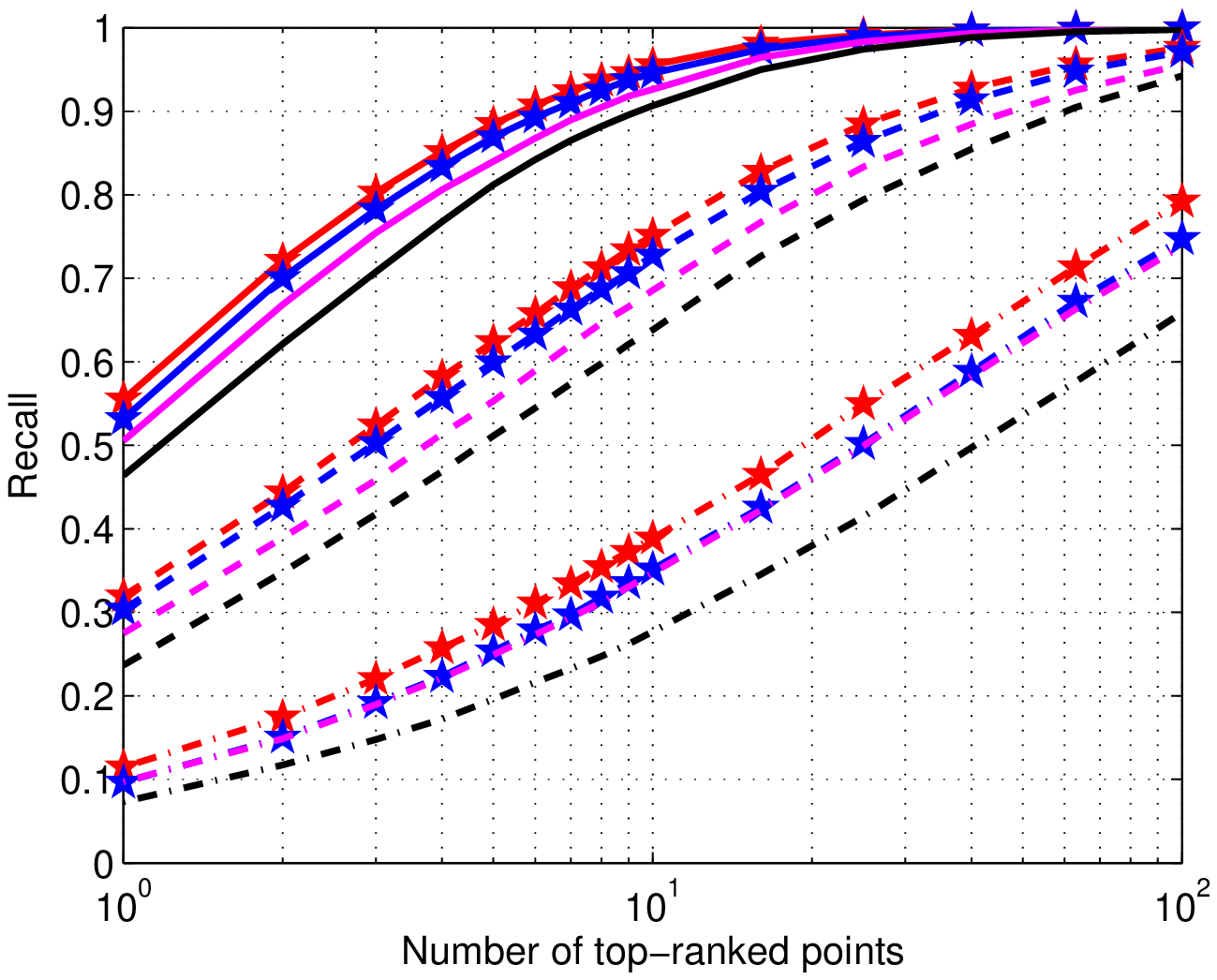} & 
\includegraphics[width = 0.3\linewidth]{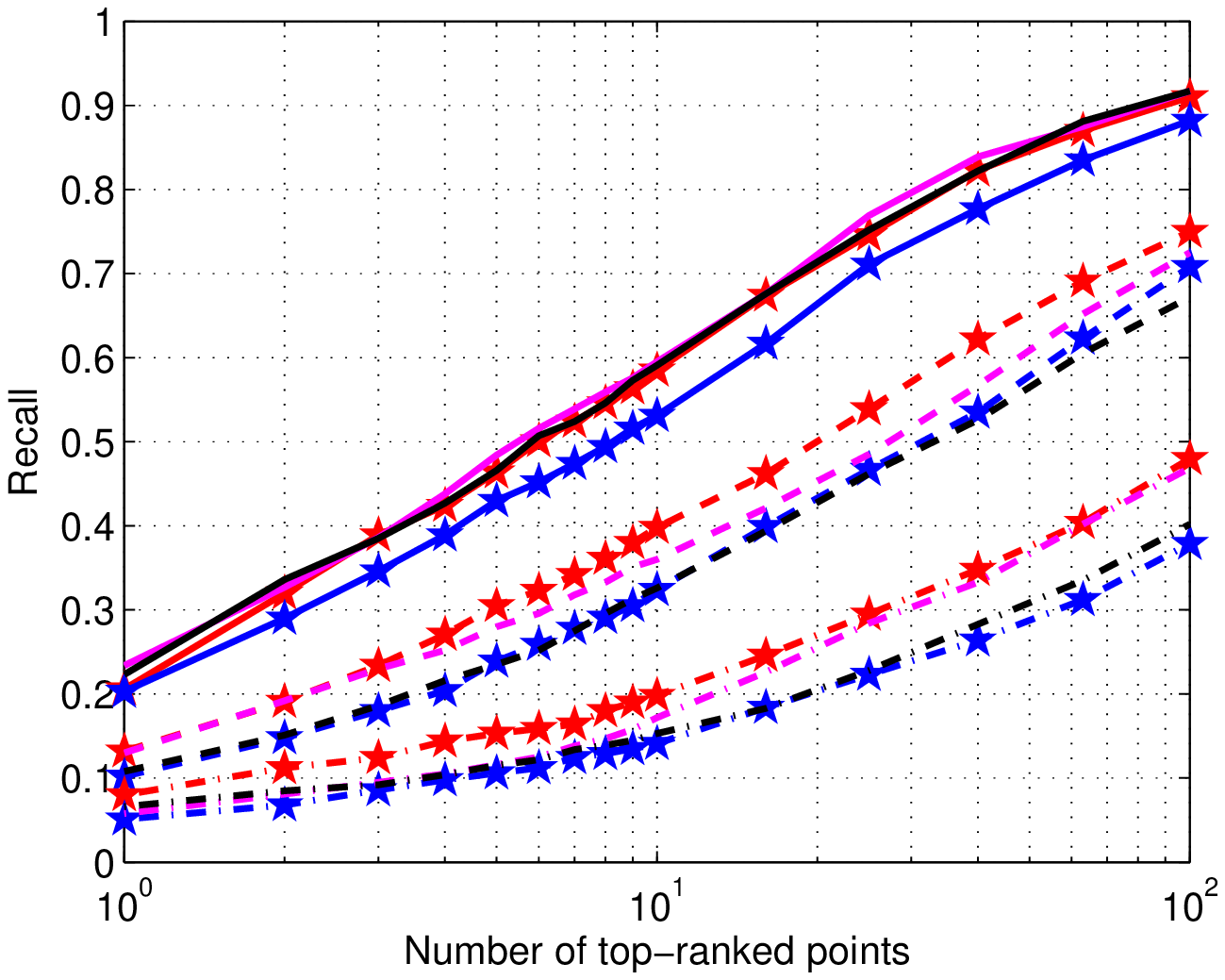} & 
\includegraphics[width = 0.3\linewidth]{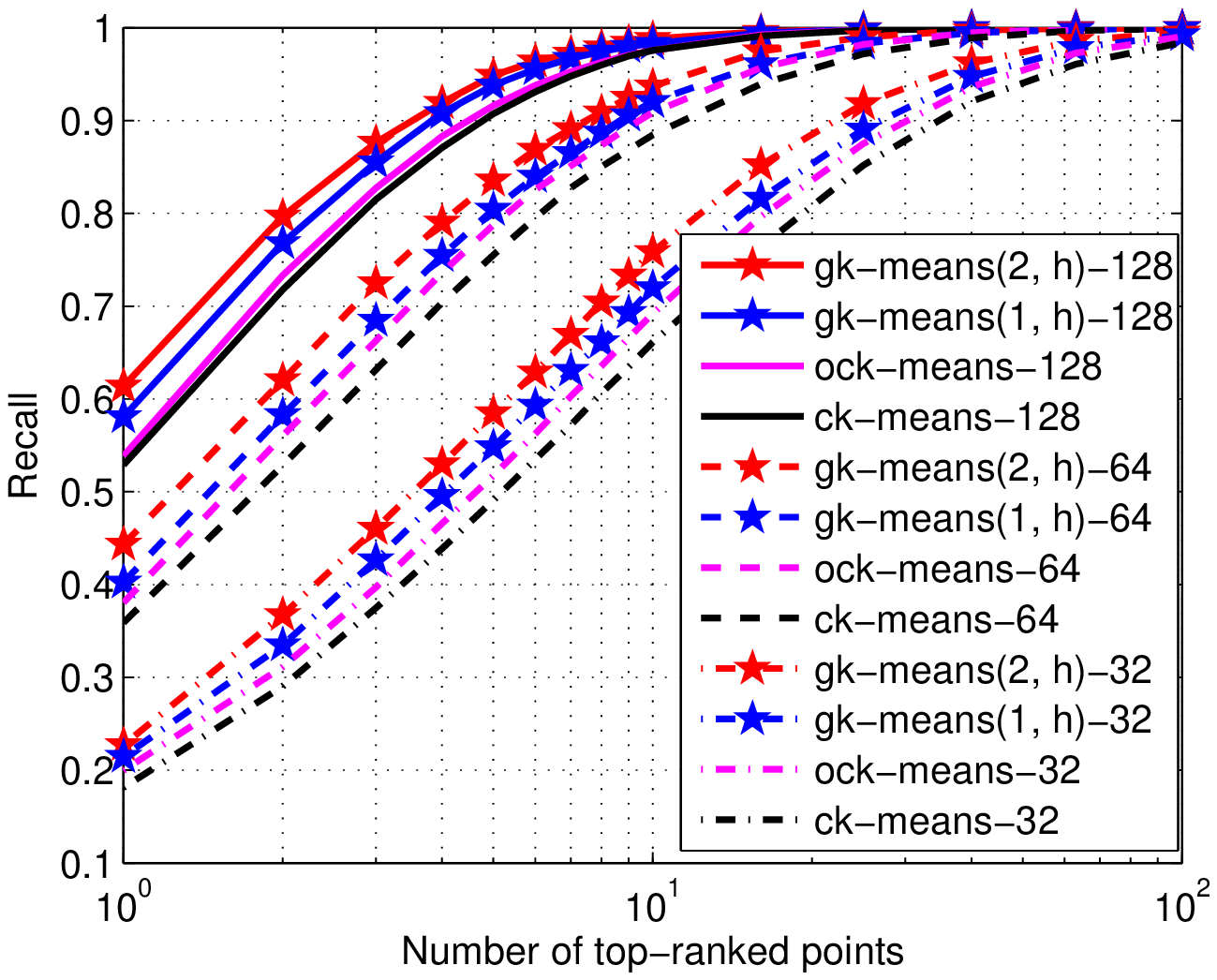} \\
(a) SIFT1M & (b) GIST1M & (c) MNIST
\end{tabular}
\caption{Recall on the three datasets with different code lengths.}
\label{figs:rec_all_code_length}
\end{figure}

The results are shown in Fig.~\ref{figs:rec_all_code_length}
on the three datasets with code length $32$, $64$, $128$.
The suffixes in the legends denote the code length. 
We can see
gk-means(2, h) almost always 
outperforms the others. 
The performance of gk-means(1, h)
is better than ock-means
on SIFT1M and MNIST, but is worse on GIST1M.
The performance varies because the data distributions among the datasets
are different and the numerical optimization
does not achieve the theoretically optimal solution.
The advantage of gk-means(2, h) over gk-means(1, h)
is because the local minimum issue of gk-means(1, h)
is more severe.
Besides, the performance comparison for ANN search is quite consistent
with the relative distortion on the database 
as shown in Table 1 in the paper.

\end{appendices}

\newpage

\bibliographystyle{plain}
\bibliography{../../../../../bib}

\end{document}